\documentclass[%
 reprint,
superscriptaddress,
nofootinbib,
 amsmath,amssymb,
 aps,
 longbibliography,
]{revtex4-1}

\usepackage[utf8]{inputenc} 
\usepackage[T1]{fontenc}    
\usepackage[colorlinks=true,urlcolor=blue,linkcolor=blue,citecolor=blue]{hyperref}       
\usepackage{url}            
\usepackage{booktabs}       
\usepackage{amsfonts}       
\usepackage{nicefrac}       
\usepackage{microtype}      
\usepackage{textgreek}
\usepackage[frozencache,cachedir=minted-cache]{minted}
\usepackage{graphicx,caption}
\usepackage{wrapfig}
\usepackage[toc,page]{appendix}
\usepackage{subcaption}
\usepackage{todonotes}
\usepackage{afterpage}

\usepackage{float}
\makeatletter
\let\newfloat\newfloat@ltx
\makeatother

\usepackage{algorithm}
\usepackage{algpseudocode}

\newcommand{\jax}{\texttt{JAX}}

\begin{document}

\preprint{APS/123-QED}

\title{Branches of a Tree: Taking Derivatives of Programs with \\Discrete and Branching Randomness in High Energy Physics}

\author{Michael Kagan}
\altaffiliation{Corresponding authors contributed equally to this work:\\ \href{mailto:makagan@slac.stanford.edu,l.heinrich@tum.de}{makagan@slac.stanford.edu,l.heinrich@tum.de}}
\affiliation{SLAC National Accelerator Laboratory}
\author{Lukas Heinrich}
\altaffiliation{Corresponding authors contributed equally to this work:\\ \href{mailto:makagan@slac.stanford.edu,l.heinrich@tum.de}{makagan@slac.stanford.edu,l.heinrich@tum.de}}
\affiliation{Technical University of Munich}

\begin{abstract}
   We propose to apply several gradient estimation techniques to enable the differentiation of programs with discrete randomness in High Energy Physics. Such programs are common in High Energy Physics due to the presence of branching processes and clustering-based analysis. Thus differentiating such programs can open the way for gradient based optimization in the context of detector design optimization, simulator tuning, or data analysis and reconstruction optimization. We discuss several possible gradient estimation strategies, including the recent Stochastic AD method, and compare them in simplified detector design experiments. In doing so we develop, to the best of our knowledge, the first fully differentiable branching program.
\end{abstract}

\maketitle

\section{Introduction}

Gradient-based optimization methods are at the core of many modern successes in Machine Learning (ML) and Artificial Intelligence (AI), especially Deep Learning. The development and application of these ML methods in High Energy Physics (HEP) has similarly enabled large performance improvements in a wide array of tasks, such as particle reconstruction, fast simulation, and parameter inference (for recent topical reviews, see e.g.~\cite{Radovic2018,10.1093/ptep/ptac097,doi:10.1142/12200,doi:10.1073/pnas.1912789117,Shlomi_2021,thais2022graph,10.21468/SciPostPhys.14.4.079,adelmann2022new}). Gradient-based optimization methods rely on automatic differentiation (AD)~\cite{autodiff,JMLR:v18:17-468}, an algorithmic way to efficiently compute the derivatives of numerical software. AD is a general tool that can be applied to scientific software beyond ML, such as simulators and inference algorithms, and used for optimizing the parameters of this software. 

The broader application of AD to numerical software, potentially mixed with ML components, is often referred to as \textit{Differentiable Programming} (DP). For instance, combining AD-enabled HEP software with ML can lead to optimizable hybrid physics-AI models with built-in physics knowledge from the HEP software, such as AI-augmented / AI-guided simulation and reconstruction, or analysis-by-synthesis inference methods with simulators in the loop~\cite{Heinrich:2022xfa, Simpson:2022suz,Cheong_2022}. Using such AD-integrated HEP software in ML models can be considered a complimentary approach to adding inductive bias into ML models through structure and architecture, such as symmetry equivariance and relational inductive bias.  

While interest is quickly growing to apply gradient-based optimization methods to a broader set challenges in HEP, such as detector design or end-to-end reconstruction, a major limitation has been the fact that standard AD can only compute derivatives of deterministic continuous functions or stochastic functions with reparametrized continuous random variables~\cite{glasserman2013,kingma2022autoencoding,10.5555/3044805.3045035}. Specifically, in HEP, many programs are both stochastic and rely on sampling \emph{discrete} random variables or decisions, such as branching points in parton showers, particle-material interactions, or clustering steps in jet building. Standard AD may not compute the desired derivative of such programs correctly, particularly when the discrete stochasticity depends on the parameter one aims to optimize (and thus differentiate with respect to). Instead, more careful consideration on how to compute the appropriate derivative is required.

There are several methods for gradient-based optimization in programs with discrete randomness, which we explore within the context of HEP applications. One method uses the \textit{score}-based approach to estimating derivatives of expectations values~\cite{williams1992}, which has been examined sparsely within HEP and not for tasks such as detector design optimization. Recently, Arya \textit{et al.}~\cite{stochAD} proposed a new AD method for handling and composing programs with discrete randomness. Using these tools, we develop simplified differentiable HEP simulators,which nonetheless exhibit the critical behaviors which until now hampered progress, and case studies to examine the behavior of and the variance of these gradient estimators. 

A review of methods for computing derivatives of stochastic programs is found in Sec.~\ref{sec:review}. Related work is discussed in Sec.~\ref{sec:related}.  Sec.~\ref{sec:app} presents applications and comparisons of different gradient estimators in HEP case studies, with an emphasis on detector design optimization. 

\vspace{0.3cm}
\noindent \textbf{Contributions\footnote{Project code can be found at \\ \url{https://github.com/lukasheinrich/branches_of_a_tree/}}:}
We introduce several methods to enable differentiation through the discrete randomness to HEP programs. We show how the score function based approach can be used for design optimization in HEP. While the score function is often used outside of HEP for design optimization, and has been used for other tasks within HEP, it has not yet been explored within the context of HEP detector design optimization. We also introduce stochastic AD~\cite{stochAD} to HEP and its ability to enable differentiable programming even in programs with discrete randomness. We provide the first application of stochastic AD to branching processes and in doing so we develop the first (to the best of our knowledge) differentiable branching program. To test these methods, we provide comparisons of gradient methods on detector design optimization case studies.

\section{Review of Differentiation of Stochastic Programs}\label{sec:review}

\begin{figure} \centering
    \begin{tabular}{cc}
    \begin{minipage}[T]{0.5\linewidth}
    \begin{minted}[escapeinside=||, fontsize=\small, linenos, numbersep=4pt, mathescape=true]{python}
def f(|$\theta$|):
    p = 0.5
    b |$\sim$| Bernoulli(p)
    return g(|$\theta$|) + b 
    \end{minted}
    \end{minipage} \hspace{-2em} \vline width .8pt & \hspace{1em}
    
    \begin{minipage}[T]{0.5\linewidth}
    \begin{minted}[escapeinside=||, fontsize=\small, linenos, numbersep=4pt, mathescape=true]{python}
def f(|$\theta$|):
    p = h(|$\theta$|)   
    b |$\sim$| Bernoulli(p)
    return g(|$\theta$|) + b 
\end{minted}
    \end{minipage} 
\end{tabular}
    \vspace{0.1cm}
    \caption{Assuming differentiable $g:\mathbb{R} \to \mathbb{R}$ and $h: \mathbb{R} \to (0, 1)$, \textbf{Left:} Toy program without $\theta$-dependence in the discrete stochastic behavior. As such, the derivative of the expected value of this program is $\frac{dE[f(\theta)]}{d\theta} = \frac{dg(\theta)}{d\theta}$ and standard AD can correctly estimate this.  
    \textbf{Right:} Toy program with $\theta$-dependence in the discrete stochastic behavior through the Bernoulli parameter $p$. Standard AD would ignore this dependence, and the resulting derivative estimator would be the same as the program on the left. However the correct derivative should be $\frac{d\mathbb{E}[f(\theta)]}{d\theta} = \frac{dg(\theta)}{d\theta} + \frac{dh(\theta)}{d\theta}$. }
    \label{fig:toy}
\end{figure}

In many HEP applications, a quantity of interest can be formulated as an expectated value of a function $f(x,\theta)$ over a parametrized density $p_\theta(x)$: $\bar{f}(\theta) = \mathbb{E}_{p_{\theta}(x)}[f(x,\theta)]$. For optimizing such quantities one requires the gradients of these expectation values of stochastic programs, e.g. $\frac{d}{d\theta} \mathbb{E}_{p_{\theta}(x)}[f(x,\theta)]$. Importantly, the expected value of such stochastic programs may be continuous and differentiable, even when they depend on discrete randomness. For instance, the expected value of a Bernoulli random variable $b\sim$Bernoulli($\theta$) has the derivative $\frac{d}{d\theta}\mathbb{E}[b] = \frac{d}{d\theta} \theta = 1$. However, standard AD tools applied to such expectations may not produce the desired result. For instance, Figure~\ref{fig:toy} shows two programs with discrete stochasticity. On the left, the discrete stochasticity does not depend on the parameter of differentiation $\theta$, and standard AD will produce the correct derivative. On the right the discrete stochasticity depends on $\theta$, standard AD will ignore this dependence and the resulting derivative will be incorrect as it will ignore this dependence. Handling these challenges requires more dedicated consideration on home to compute the appropriate derivative. 
We briefly review several approaches to gradient estimation below (see Ref.~\cite{mohamed2020monte} for a detail review).

\vspace{0.4cm} \noindent
\textbf{Finite Differences (Numerical Differentiation)}:
Finite difference methods estimate derivatives by computing the difference between forward evaluations of a program and a perturbed version of the program, for instance:

\begin{equation}
    \frac{d}{d\theta} \mathbb{E}_{p_{\theta}(x)}[f(x)] \approx \frac{\mathbb{E}_{p_{\theta+\epsilon}(x)}[f(x)] - \mathbb{E}_{p_{\theta}(x)}[f (x)]}{\epsilon}
\end{equation}

Finite difference methods are prone to high variance~\cite{JMLR:v18:17-468}, and require large numbers of program evaluations when $\theta$ is high dimensional. Central difference methods can reduce error. One large contributor to this large variance is that multiple independent evaluations of the program are used to estimate this gradient, introducing separate stochastic evaluation paths of the program.

\vspace{0.4cm} \noindent
\textbf{Reparameterization Trick}:
In many cases, when sampling $x \sim p_{\theta}(x)$, we can \textit{smoothly reparametrize} this sampling as $z \sim p(z)$ and $x = g(z, \theta)$, where $p(z)$ is often a simple ``base" distribution and $g(\cdot, \theta)$ provides a differentiable, $\theta$-dependent transformation of samples from the base to the desired distribution. For example, the normal distribution $x \sim \mathcal{N}(\mu, \sigma)$ is location-scale reparameterizable through $z \sim \mathcal{N}(0,1)$ and $x = \sigma z + \mu \sim \mathcal{N}(\mu, \sigma)$. This is particularly convenient for computing derivatives of expectation values of differentiable functions $f(\cdot)$, 
\begin{eqnarray}\nonumber
    \frac{d}{d\theta} \mathbb{E}_{p_{\theta}(x)}[f(x)]  &=& \frac{d}{d\theta} \int p(z) f(g(z, \theta)) dz \\
     &=& \int p(z) \frac{df}{dg} \frac{dg(z, \theta)}{d\theta}dz
\end{eqnarray}
Many HEP simulators, which implement a structural causal model, can be considered as a reparametrization, mapping from noise variables to random variables $x$ with physical meaning. However if the random variables $x$ are discrete, the map is not differentiable, which limits the applicability of the reparametrization trick within a HEP context.

\vspace{0.4cm} \noindent
\textbf{Surrogate Methods}:
When a reparameterization is not possible, either because $f(\cdot)$ is non differentiable or $p_{\theta}(x)$ does not admit a smooth reparameterization, surrogate methods can be used. In this case, an ML model $S(z, \theta)$, with $z \sim p(z)$ a chosen distribution, is trained to mimic the stochastic program. As such, surrogate methods try enable reparameterization though a ML model and thus enable differentiation, for instance:

\begin{eqnarray}\nonumber
    \frac{d}{d\theta} \mathbb{E}_{p_{\theta}(x)}[f(x)] &\approx& \frac{d}{d\theta} \int p(z) S(z, \theta) dz  \\
     &=& \int p(z) \frac{dS(z, \theta)}{d\theta}dz
\end{eqnarray}

The quality of this derivative estimator will depend significantly on the quality of the surrogate model as an approximation of the original program. Moreover, the surrogate will learn a smooth approximation of non-differentiable elements of the program, but how this approximation is performed and if bias is introduced is difficult to asses.

\vspace{0.4cm} \noindent
\textbf{Score function}:
When the parameter dependence of a differentiable distribution $p_{\theta}(x)$ is known and differentiable with respect to the parameters, one can compute:
\begin{eqnarray}\label{eq:score} \nonumber
\frac{d}{d\theta} \mathbb{E}_{p_{\theta}(x)}[f(x)] &=&  \int p_{\theta}(x) \frac{d \log p_{\theta}(x)}{d\theta}  f(x) dx \\
&=& \mathbb{E}_{p_{\theta}(x)}\Big[\frac{d \log p_{\theta}(x)}{d\theta}  f(x)\Big]
\end{eqnarray}
where $\frac{d}{d\theta} \log p_{\theta}(x)$ is known as the score function. This gradient estimator is also known as REINFORCE~\cite{williams1992}, and is commonly used in reinforcement learning. The benefit of this approach is that the $f(x)$ does not need to be differentiable, only $p_{\theta}(x)$ must be differentiable with respect to $\theta$. As such, discrete random variables can be used in the computation of $f(x)$.

\vspace{0.3cm} \noindent
\textit{Control Variates}: Score function based gradient estimates often have a large variance, which can make tasks like optimization with gradient descent slow and more difficult. A control variate, $c(x, \theta)$, can be subtracted from $f(x)$ in Eqn.~\ref{eq:score} to reduce the variance of the estimator as long as it does not bias the estimator, i.e. as long as $\mathbb{E}_{p_{\theta}(x)}\Big[\frac{d \log p_{\theta}(x)}{d\theta}  c(x, \theta)\Big] = 0$. Noting that $\mathbb{E}_{p_{\theta}(x)}\Big[\frac{d \log p_{\theta}(x)}{d\theta} \Big] = 0$, one way to find a control variate  is to choose a $c(\theta)$ which does not depend on $x$. A common control variate, often also called a baseline, is the function mean $\bar{f}_{\theta} = \int p_\theta (x) f(x)$; in practice $\bar{f}_{\theta}$ is often estimated using the mean of a mini-batch.  We will use this baseline for the experiments in Sec.~\ref{sec:app}. More details on variance reduction methods for score based gradient estimation can be found in Ref.~\cite{10.5555/1005332.1044710}.

\vspace{0.3cm} \noindent
\textit{Proposal Distributions}: 
In some case, we may not know or have access to $p_{\theta}(x)$, for example when $g(\theta)$ is a simulator with parameters as input and sampling is done internally within the program. 
One approach can be to imagine the input to the program as a sample from a proposal distribution $\theta \sim \pi_{\psi}(\theta)$, where $\psi$ are the parameters of the proposal distribution. For instance one could choose a normal distribution $N(\psi, 1)$ for the proposal. One would then aim to optimize the mean of this proposal,
\begin{equation}
\frac{d}{d\psi} \mathbb{E}_{\pi_{\psi}(\theta)}[g(\theta)] = \mathbb{E}_{\pi_{\psi}}\Big[\frac{d \log \pi_{\psi}(\theta)}{d\psi} g(\theta)\Big]
\end{equation}

\vspace{0.4cm} \noindent
\textbf{Stochastic AD}:
The stochastic derivative of the expected value of a function $f(\cdot)$ of a discrete random variable $x \sim p_\theta(x)$ has the form~\cite{stochAD}: 

\begin{equation}\label{eq:stochad}
\frac{d}{d\theta} \mathbb{E}_{p_\theta(x)}[f(x)] = \mathbb{E}_{p_\theta(x,y)}[\delta + \beta \big(f(y) - f(x)\big) ]
\end{equation}
where $\delta$ is the standard derivative $\partial f / \partial \theta$ as computed with AD, and the second term corresponds to the effect of a change in $\theta$ on the sampling of the discrete random variable $x$. The weight $\beta$ depends on the underlying sampling distribution and $y$ is an alternative value of the discrete random variable. Conceptually, for discrete $X$ with consecutive integer range one can understand this result through the lens of reparameterization.  One can reparamaterize the discrete distribution via the inversion method, e.g.:
\begin{eqnarray}
    \nonumber \omega &\sim& U[0,1] \\
    \nonumber x &=& \left\{ X\ : \omega \in [CDF_{p_\theta}(X), CDF_{p_\theta}(X+1)\ )\ \right\}
\end{eqnarray} 
For example, if $x$ is a Bernoulli random variable with parameter $\theta$, then one can reparameterize $x = H(\omega > 1-\theta)$ where $H(\cdot)$ is the Heaviside step function. The boundaries which define the set of values of $\omega$ that result in a value of $X$ are now dependent on the parameters $\theta$. A change in parameters changes the boundaries, and thus changes the  probabilities of different outcomes $x$. The second term of the stochastic AD derivative accounts for the infinitesimal change in probabilities as the boundaries are changed as well as the alternative value of the program $y$ that would result from such a change.
Importantly, one can define this derivative at each program evaluation and at each stochastic sampling within the program, allowing for the development of composition rules and of forward mode automatic differentiation.  
For a more detailed discussion of Stochastic AD, see Ref.~\cite{stochAD}.

The expectation on the right hand side of Equation~\ref{eq:stochad} is taken with respect to the joint distribution $p_\theta(x,y)$, which is often denoted the \textit{coupling}. The marginal distributions of this coupling must be the same as the original sampling distribution, e.g. $\int dy p_\theta(x,y) = \int dx p_\theta(x,y) = p_\theta (x)$ to ensure both the primal evaluation of the program and the alternative proceed under the normal operation of the program. As such, there is a distribution over possible alternative programs.  In practice, out of all possible alternatives from all of the discrete samplings within a program, only one alternative is randomly chosen. This \textit{pruning} process treats the set of alternatives as a categorical distribution with the probability of a given alternative being the weight of the alternative relative to the total event weight (computed using the composition rules). This alternative may occur in the middle of the program, and is then tracked in parallel to the primal until completion of the alternative program. One can then use this program alternative $y$ for computing even gradients using Equation~\ref{eq:stochad}.

\vspace{0.3cm} \noindent
\textit{Variance Reduction with Random Number Reuse}: While the marginals of the coupling are fixed, one has considerable flexibility to choose the correlation structure between $x$ and $y$. This is important because once an alternative is determined from a discrete sampling within the program, the alternative program will be run to completion and thus may require additional sampling of discrete random variables. The more correlated are the evaluations of the alternative completion to that of the primal evaluation, the lower the variance of the gradient estimator may be. As downstream discrete samplings also occur in the primal program, one can reuse the reparameterized sampling in the primal program, i.e. the $\omega$ values sampled for the inversion method. By reusing $\omega$ values, less additional variance is added to the alternative program then if the downstream discrete random variable were sampled independently. In this work for the experiments in Sec.~\ref{sec:app}, we use a first-in first-out (FIFO) approach, where at each time step we store $\omega$ values in the FIFO while iterating over branches (particle) in the primal, and then pull $\omega$ values from the FIFO while iterating over branches in the alternative. If additional $\omega$ values are needed by the alternative, they are sampled independently of the primal.

\section{Related Work}\label{sec:related}

Automatic differentiation~\cite{autodiff,JMLR:v18:17-468}, and its use in gradient based optimization, is ubiquitous in ML, statistics, and applied math. AD uses the chain rule to evaluate derivatives of a function that is represented as a computer program.  AD takes as input program code and produces new code for evaluating the program and derivatives. AD typically builds a computational graph, or a directed acyclic graph of mathematical operations applied to an input. Gradients are defined for each operation of the graph, and the total gradient can be evaluated from input to output, called forward mode, and from output to input, called reverse mode or backpropagation in ML. AD is backbone of ML / DP frameworks like \texttt{TensorFlow}~\cite{tensorflow2015-whitepaper}, \jax~\cite{jax2018github}, and \texttt{PyTorch}~\cite{NEURIPS2019_9015pytorch}. 

Significant work has been performed on developing Monte Carlo estimators for gradients in machine learning, as discussed in the recent review~\cite{mohamed2020monte}, and gradients of stochastic computation graphs~\cite{NIPS2015_de03beff}. More recently methods such as Stochastic AD~\cite{stochAD,arya2023differentiating} have been developed to target  derivatives of programs with discrete stochastic behavior in a compositional way, as well as developing specific applications with dedicated variance reduction methods (e.g. developing coupling for these applications).   
Extensions to AD have recently been proposed for differentiating the expectations of stochastic programs~\cite{Lew_2023,NEURIPS2021_3dfe2f63}, and to account for parametric discontinuities~\cite{BangaruMichel2021DiscontinuousAutodiff}.

Differentiable programming approaches have begun to be explored in HEP. Examples include histogram fitting in \texttt{pyHF}~\cite{pyhf, pyhf_joss}, in analysis optimization in \texttt{Neos}~\cite{Simpson:2022suz,lukas_heinrich_2020_3697981}, for modeling parton distribution functions (pdf) used by matrix element generators~\cite{pdflow2021,ball2021opensource}, and for developing differentiable matrix element simulations with \texttt{MadJax}~\cite{Heinrich:2022xfa}. Related to our work, Ref.~\cite{Nachman:2022jbj} studies differentiating a parton shower, which focuses on the derivative of the primal shower program but does not examine differentiation through discrete randomness in such branching programs. Within cosmology, the Differentiable Universe Initiative is developing a differentiable simulation and analysis toolchain for comological data analyses~\cite{DiffUniv}.

Within HEP, score function gradient estimators have been used within the context of jet grooming~\cite{Carrazza:2019efs} and hierarchical jet clustering~\cite{brehmer2020hierarchical}. To the best of our knowledge, this work is the first application to detector design optimization in HEP.

 On HEP detector design, surrogate based optimization methods have been developed and explored for particle detectors in Ref.~\cite{NEURIPS2020_a878dbeb}. Surrogate methods have also been applied to neutrino detector design~\cite{chaack_icrc}. Detector design optimization with standard AD tools and with surrogate methods is discussed in Ref.~\cite{MODE:2022znx}. A branch-and-bound type algorithm is explored in Ref.~\cite{Gorordo:2022rro}.

\begin{figure*}[ht!]
\centering

    \begin{subfigure}[b]{0.4\textwidth}
    \centering
    \includegraphics[width=0.94\textwidth]{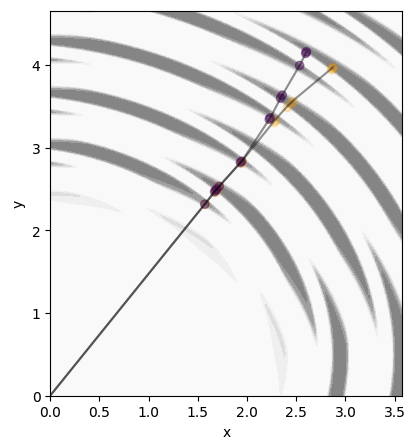}
    \caption{Single particle energy loss}
    \label{fig:event_display_single}
    \end{subfigure}
    \begin{subfigure}[b]{0.4\textwidth}
    \centering
    \includegraphics[width=\textwidth]{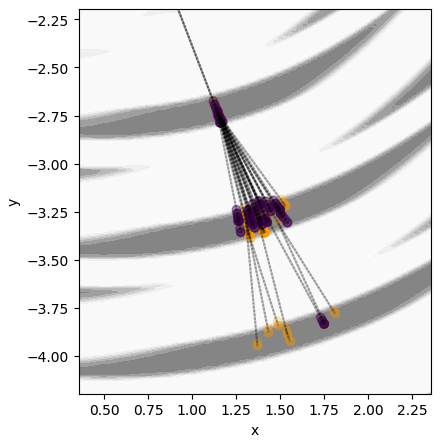}
    \caption{Particle shower with splitting}
    \label{fig:event_display_shower}
    \end{subfigure}

\caption{Event Displays of a single particle energy loss (left) and a particle shower with particle splitting (right). A primal event is shown in purple, while an alternative event, as determined using Stochastic AD, is shown in yellow. The material map is shown in grey.}
\label{fig:event_display}
\end{figure*}

\section{Applications}\label{sec:app}

We present a series of applications in a simplified simulation of particles interacting with detector material. In these experiments, we examine how different gradient estimation methods can be applied and compare their performance in terms of the variance of the estimators.

\subsection{Particle Interaction Simulator}

The simplified simulator in this work aims to capture the salient features of particle physics simulators that model the traversal of particles through matter but be simple enough allow reimplementation in a programming language of choice for a detailed study of various gradient estimation in a self-contained setting.

\begin{algorithm}[ht!]
  \caption{Simplified Particle Interaction Simulator}
  \label{alg:train}
  \begin{algorithmic}[1]
    \Require $E_0$ : energy threshold
    \Require $\epsilon$ : energy loss at interaction
    \Require $m(x,\theta): \mathbb{R}^3 \times \mathbb{R}^n \to [0,1]$ : material map
    \Require $P = \{p_i = (\vec{x}_i \vec{p}_i,E_i)\}$: initial particle list
    \Require $H = \emptyset$: list of hits
    \Function{Simulate}{$P$,$\theta$}
    \While{not all $p\in P$ below threshold}
        \State $P \leftarrow \emptyset$ \Comment{list of surviving particles}
        \For{for all particles in $P$}

            \If {$E_i < E_0$}
                \State continue  \Comment{particle below thr.} 
            \EndIf

            \State $p_i = \mathrm{propagate}(p_i)$
            \State $\rho_\mathrm{E} \leftarrow$  \verb+sample+ $\mathrm{Ber}(\rho_\mathrm{E}|m_\theta(x_i))$
            \If {$\rho_E$} \Comment{particle interacts}
                \State $H \leftarrow H \cup \{x_i\}$ \Comment{add position to hits}

                \State $\rho_\mathrm{split} \leftarrow$  \verb+sample+ $\mathrm{Ber}(\rho_\mathrm{split}|m_\theta(x_i))$
                \If {$\rho_\mathrm{split}$} \Comment{particle splits}
                    \State $p_L,p_R \leftarrow \mathrm{split}(p_i)$
                    \State $P \leftarrow P \cup \{p_R, p_L\}$
                \Else
                    \State $E_i \leftarrow E_i - \epsilon$ \Comment{lose energy}
                    \State $P \leftarrow P \cup \{p_i\}$
                \EndIf
            \Else
                \State $P \leftarrow P \cup \{p_i\}$
            \EndIf
      \EndFor
    \EndWhile\label{euclidendwhile}
    \State \textbf{return} $H$
    \EndFunction
  \end{algorithmic}\label{alg:simulate}
\end{algorithm}

The simulator $S_\theta(p)$ models the stochastic evolution of particles through two main processes. A binary \emph{splitting process} $p_0 \to p_L, p_R$ that splits a parent particle momentum evenly across two child particles and a \emph{energy loss process} $E\to E-\epsilon$. The probability of a particle interaction is modeled as a function of the material map $m(x)$. The simulation is performed by fixed time steps, first propagating particles in their direction of travel, and then querying the material map to determine if an interaction occurs, and if so which type of interaction (i.e. splitting or energy loss). Pseudo-code for the simulator can be found in Alg.~\ref{alg:simulate}.

The detector is simulated as a continuous material map $m_{\theta}(x_0, x_1)$ which takes as input a position $(x_0,x_1)$ and outputs an interaction probability. This interaction probability is dependent on detector parameters $\theta$, and we will examine examples where derivatives with respect to $\theta$ are sought. Only a single detector parameter is used in the following experiments, which is the detector inner radius which will be denoted $\theta_R$. 
With $r=\sqrt{x_0^2 + x_1^2}$, and $\phi = \arctan{\frac{x_0}{x_1}} $, the material map is defined as:
\begin{equation}
m_\theta(x,y) = \frac{1}{2} m_{start}(r, \theta_R)\  m_{\Phi}(r, \phi)\ m_{R}(r)\ m_{end}(r, \theta_R)
\end{equation}
where
\begin{eqnarray} 
    \nonumber m_{start}(r, \theta_R) &=& \frac{1}{1+e^{-\beta(r-\theta_R)}} \\ 
    \nonumber m_{\Phi}(r, \phi) &=& \frac{1}{1+e^{\beta\sin(\omega (\phi + 2r))}} \\ 
    \nonumber m_{R}(r) &=& \frac{1}{1+e^{\beta\cos(\omega (r-2))}} \\ 
    \nonumber m_{end}(r, \theta_R) &=& \frac{1}{1+e^{\beta(r-\theta_R - R_{max})}} \ .
\end{eqnarray}

The terms $m_{start}(r, \theta_R)$ and $m_{end}(r, \theta_R)$ determine the inner and outer radius of the detector, respectively. The terms $m_{\Phi}(r, \phi)$ and $m_{R}(r)$ determine the segmentation in $\phi$ and $r$, respectively.
The constants $\beta$, $\omega$, and $R_{max}$ control the sharpness of the smooth material map, the segmentation frequency in the azimuthal direction, and the maximum depth of the detector, respectively. The parameter that will be optimized is $\theta_R$, the inner radius of the detector.

Example material maps can be seen in grey in the event displays of Figure~\ref{fig:event_display}, where the darkness of the shade of grey indicates the strength of interaction.

\subsection{Single Particle Energy Loss}

\begin{figure*}[ht!]
\centering
    \centering
    \includegraphics[width=\textwidth]{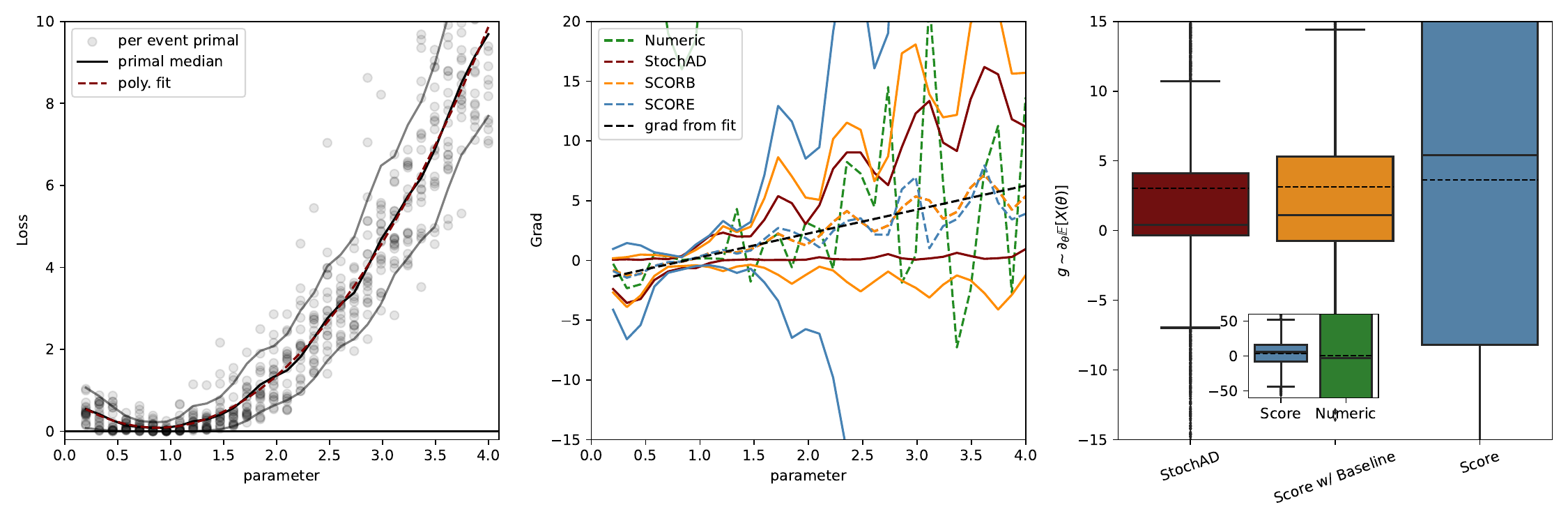}

\caption{For simulations of single particle energy loss: (\textit{Left}) The loss function and various gradient estimators of the loss are shown as a function of the detector radius parameter. Sample primal evaluations of the loss are shown as markers, and the interquantile interval is shown in black. The red dashed line shows the derivative of an polynomial interpolation of the mean loss. 
(\textit{Middle}) The mean and standard deviation of the numeric, Stochastic AD (STAD), score function (SCORE), and score function with baseline (SCORB) gradient estimators as a function of the detector radius parameter. The gradient of the polynomial interpolation of the mean loss is shown in dashed black. (\textit{Right}) Box plot of the four variance estimators evaluated at parameter value $\theta_R=2.5$m, with the mean shown as a dashed line.}
\label{fig:eloss_summary}
\end{figure*}

In the single particle energy loss setting, interactions which cause splitting are turned off, i.e. there are no showers. At each step, the particle interacts with the detector with a probability $p_\mathrm{Eloss} = m(x,\theta)$ which is dependent on material map parameters $\theta$. This probability is large is high density regions of detector material and small in low density regions. A Bernoulli distribution with parameter $m(x,\theta)$ is sampled at each time step to determine if the interaction occurs and, if so, the particle deterministically loses energy $\epsilon = 1$ GeV. All particles are set to have initial energy of 25 GeV and when the particle energy falls below $E_0=0.5~\mathrm{GeV}$, the particle is stopped. An example event display can be seen in Fig.~\ref{fig:event_display_single}, where the primal particle trajectory is seen in purple, the alternative trajectory determined with Stochastic AD is seen in yellow, and the material map is seen in grey.  

In this example, there is only one detector parameter $\theta \equiv R$, the inner radius of the detector, and derivatives are computed with respect to $R$ using several methods of estimating gradients. The mean squared error between the radial position of points of particle interaction and a target radius $\bar{R}_T$ is used as a loss function that one may minimize for the purposes of design optimization. In this example we set $\bar{R}_T = 2$m. 

The loss as a function of the detector radius parameter can be seen on the left in Fig.~\ref{fig:eloss_summary}. The loss from individual primal simulation samples can be seen in grey, the median loss and interquartile range in black, and a polynomial interpolation of the average loss. Even though the simulation is stochastic, and one can see the variations of the loss at each parameter in the grey points, there is a clear minimum to the loss function.

The gradient estimators and their standard deviations, calculated over the 5000 simulation runs, can be seen as a function of the detector radius parameter in the middle in Fig.~\ref{fig:eloss_summary}. The distributions of gradient estimators evaluated at the parameter value $\theta_R=2.5$m can be seen in the box plot on the right in Fig.~\ref{fig:eloss_summary}.  As expected, numerical derivatives have the largest standard deviation, though it should be noted that this can depend highly on the size of the finite different $\epsilon$ and the method for calculating the numerical derivative. Similarly, score function gradient estimators without baseline shows a high standard deviation, especially at large radius parameter where the loss function has larger standard deviation over different simulation samples. The score with baseline has a much reduced standard deviation over the score function without baseline across all parameter values. Stochastic AD shows the smallest standard deviation of all estimators, likely owing to the ability to couple much of the alternative program evaluation to that of the primal up to the alternative branching point. Across all gradient estimators, the mean of the gradient estimator, shown in solid lines, are close to the gradient of the polynomial fit of the loss  (which serves as a rough guide to the gradient of the expected loss) within one standard deviation. A numerical comparison of gradient estimator mean and variance can be found in Tab.~\ref{tab:table_gradients}.

\subsection{Branching Shower}
\begin{figure*}[t!]
\centering
    \includegraphics[width=\textwidth]{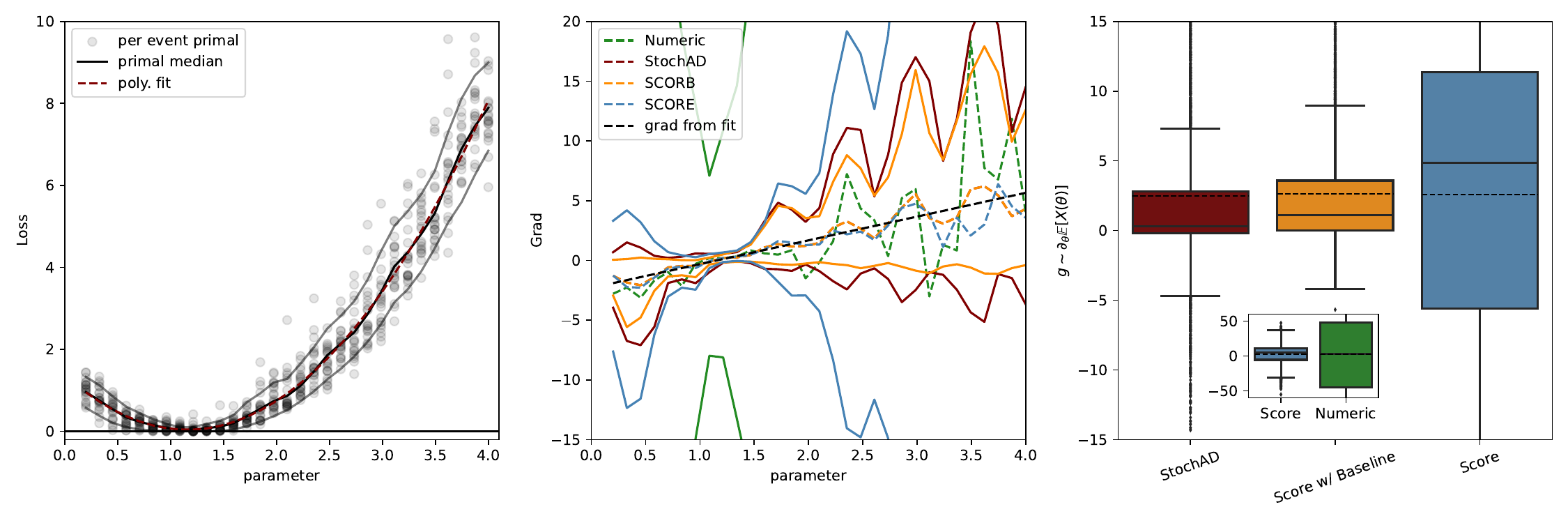}
    
\caption{For simulations of particle showers with splitting: (\textit{Left}) The loss function and various gradient estimators of the loss are shown as a function of the detector radius parameter. Sample primal evaluations of the loss are shown as markers, and the interquantile interval is shown in black. The red dashed line shows the derivative of an polynomial interpolation of the mean loss. 
(\textit{Middle}) The mean and standard deviation of the numeric, Stochastic AD (STAD), score function (SCORE), and score function with baseline (SCORB) gradient estimators as a function of the detector radius parameter. The gradient of the polynomial interpolation of the mean loss is shown in dashed black. (\textit{Right}) Box plot of the four variance estimators evaluated at parameter value $\theta_R=2.5$m, with the mean shown as a dashed line.}
\label{fig:split_summary}
\end{figure*}

In the branching shower example, the same material map, and thus interaction probability, as the single particle energy loss simulation is used but if an interaction occurs, the particle is deterministically split into two daughter particles each with half the energy of the parent particle and with an opening angle of 0.1 radians.  Initial particles are set to have starting energy of 25 GeV and when any particle energy falls below 0.5 GeV, the particle is stopped. The same loss function as in the single particle energy loss example is used here.  An example event display can be seen in Fig.~\ref{fig:event_display_shower}, where the primal particle shower is seen in purple, the alternative shower determined with Stochastic AD is seen in yellow, and the material map is seen in grey.  

The loss function and the standard deviation of the gradients, as functions of the detector radius parameter, can be seen on the left and middle, respectively, in Fig.~\ref{fig:split_summary}. As in the single particle energy loss example, the numerical gradients are found to have the largest standard deviation of gradients, with the score function without baseline estimator having the second largest standard deviation. Notably, the score function with baseline estimator is found to have the smallest standard deviation, slightly smaller than the Stochastic AD gradient estimator. Unlike the single particle example, the splitting shower has many program branching points which can create alternative outputs that are significantly different from the primal shower. In turn, this leads to a reduction in the correlation between the primal and alternative showers and ultimately to an increase in the gradient estimator standard deviation.  A comparison of the distribution of gradient estimators, at detector parameter value $\theta_R=2.5$m, can be seen on the right in Fig.~\ref{fig:split_summary}. While the mean values (dotted lines) in each box agree well across estimators, the variance estimates as well as the tails are significantly better behaved for Stochastic AD and score function with baseline estimators. Similarly, a comparison of the mean and standard deviation of the gradient estimators evaluated at the parameter value $\theta_R=2.5$m for both the single particle energy loss and the splitting shower can be found in Tab.~\ref{tab:table_gradients}.

\begin{table}[ht]
    \centering
    \begin{tabular}{lll}
        \toprule
        \textbf{Estimator} & $E$-loss & Shower\\
        \midrule
        StochAD & $\mathbf{3.17 \pm 4.47}$  &  $2.53 \pm 6.37$\\
        Score w/ Baseline & $3.01 \pm 6.59$ & $\mathbf{2.47 \pm 4.42}$\\
        Score w/o Baseline & $2.68 \pm 17.18$ & $2.76 \pm 12.20$\\
        Numerical & $3.83 \pm 139.96$ & $2.43 \pm 74.85$\\
        \bottomrule
    \end{tabular}
    \caption{Gradient estimator mean and standard deviation, for both the single particle energy loss and splitting shower, evaluated at parameter value $\theta_R = 2.5$m and determined from 5,000 samples. The estimator with lowest standard deviation is shown in bold.}
    \label{tab:table_gradients}
\end{table}

It should be noted that there is considerable flexibility in Stochastic AD for how to couple the randomness in the primal and alternative programs after the point at which the alternative is produced, i.e. how to choose the join distribution over random variables in the primal and alternative programs. 
This selection of coupling can have a considerable impact on the Stochastic AD gradient estimator variance. In this work, we have used a simple approach of re-using random variables sampled in the primal for the alternative, without regard for where those random variables are re-used in the alternative. We have seen that this re-use can have a large impact; we observed that removing the re-use of random variables in the alternative can increase the Stochastic AD gradient estimator standard deviation by factors of 1.5 or more. More generally, a more careful strategy of re-using of random variables may considerably reduce the Stochastic AD gradient estimator variance. 

\subsection{Design Optimization with Splitting Shower}

\begin{figure}[ht]
\centering

    \begin{subfigure}[b]{0.49\textwidth}
    \centering
    \includegraphics[width=\textwidth]{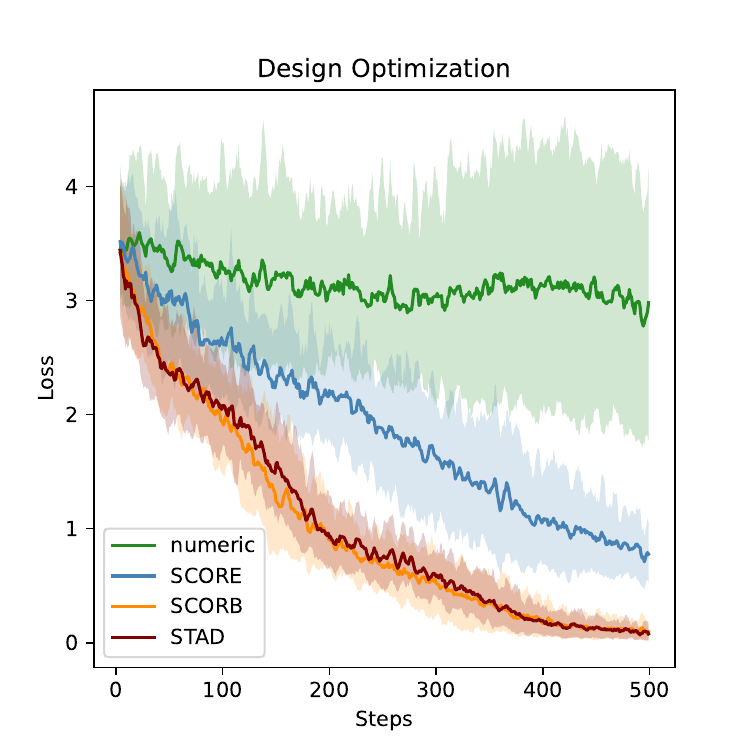}
    \end{subfigure}
    
\caption{The mean and inter-quantile range of the loss versus epoch of detector design optimization is shown. Mean and quantiles  are computed from 10 optimizations.}
\label{fig:optim_split}
\end{figure}

We test the ability to use the various gradient estimators to perform a gradient based optimization using of the detector radius parameter using the aforementioned loss with a target radial shower depth of $\bar{R}_T = 2$. Each epoch  consists of a single step of the optimization, with a mini-batch size of only 2 simulation runs used to estimate gradients in each epoch. The Adam optimizer~\cite{DBLP:journals/corr/KingmaB14} is used. A learning rate of 0.01 is used for the gradient descent parameter update. For all optimizations, the initial detector radius parameter value is  set to $\theta_{init}=3$m and the optimization is run for 500 gradient steps. Each gradient method is used in 10 separate optimizations, and the average and standard deviation of the loss at each optimization step is shown in Fig.~\ref{fig:optim_split}. As expected, the score function estimator without baseline and the numeric gradients shows the largest standard deviation to the extent that optimization is not feasible in this setting. We also see that the numeric and score function without baseline estimators are significantly slower at optimizing the objective. The score with baseline estimator and Stochastic AD estimators show similar standard deviation and similar progress towards the loss minimum as a function of optimization step.  This suggests that Stochastic AD and score function with baseline estimators provide significantly better gradient estimates, even with very small sample sizes, and are likely interesting estimators for further study of detector design optimizations in more complex settings.

\section{Conclusion}

In this work, we discuss several strategies for differentiating stochastic programs, with a focus on methods capable of differentiating programs with discrete randomness, and discuss their application to High Energy Physics detector design optimization. We develop the first application of Stochastic AD to branching processes and, more generally, the first differentiable branching program capable of estimating gradients through the discrete processes within a particle shower. We also introduce score function gradient estimators within this HEP detector design context. We find that Stochastic AD and score function gradient estimators, using control variates, provide the best gradient estimators in terms of smallest standard deviation among the gradient estimators examined within a case study of detector design. We show that both techniques can successfully be used for gradient-based HEP detector design on a toy detector simulator.

More broadly, we believe that the careful study and application of techniques like Stochastic AD and score function estimation can open the way to a wide array of new differentiable programming applications in HEP and other sciences.


\section*{Acknowledgements}
We thank Gaurav Arya, Frank Sch{\"a}fer, and Moritz Schauer for the helpful discussions regarding Stochastic AD, and thank Gaurav Arya for the helpful feedback on the manuscript. We thank Michael Brenner for the helpful discussions regarding score function gradient estimators at the Aspen Center for Physics, as this work was partially performed at the Aspen Center for Physics, which is supported by National Science Foundation grant PHY-2210452. We also thank the Munich Institute for Astro-, Particle and BioPhysics (MIAPbP) which is funded by the Deutsche Forschungsgemeinschaft (DFG, German Research Foundation) under Germany´s Excellence Strategy – EXC-2094 – 390783311, as this work was partially performed at the MIAPbP workshop on Differentiable and Probabilistic Programming for Fundamental Physics.

MK is supported by the US Department of Energy (DOE) under grant DE-AC02-76SF00515. LH is supported by the Excellence Cluster ORIGINS, which is funded by the Deutsche Forschungsgemeinschaft (DFG, German Research Foundation) under Germany’s Excellence Strategy - EXC-2094-390783311.

\bibliography{bibliography}

\end{document}